\documentclass[10pt,twocolumn,letterpaper]{article}

\usepackage{cvpr}              
\usepackage{multirow}

\usepackage[dvipsnames]{xcolor}

\definecolor{cvprblue}{rgb}{0.21,0.49,0.74}
\usepackage[pagebackref,breaklinks,colorlinks,citecolor=cvprblue]{hyperref}

\title{One-Shot Learning for Pose-Guided Person Image Synthesis in the Wild}

\author{
  Dongqi Fan\textsuperscript{1}\quad Tao Chen\textsuperscript{2}\quad Mingjie Wang\textsuperscript{3}\quad Rui Ma\textsuperscript{4}\quad Qiang Tang\textsuperscript{5}\quad \\Zili Yi\textsuperscript{6}\quad Qian Wang\textsuperscript{7}\quad Liang Chang\textsuperscript{1} \\
  \textsuperscript{1}University of Electronic Science and Technology of China \\
  \textsuperscript{2}China Mobile Zijin (Jiangsu) Innovation Research Institute Co., Ltd\\
  \textsuperscript{3}Zhejiang Sci-Tech University\quad \textsuperscript{4}Jilin University\quad \textsuperscript{5}Huawei Canada\\ \textsuperscript{6}Nanjing University\quad 
  \textsuperscript{7}China Mobile Research Institute \\
}

\begin{document}
\maketitle

\begin{figure*}[htbp]
\centering
\includegraphics[width=0.92\textwidth]{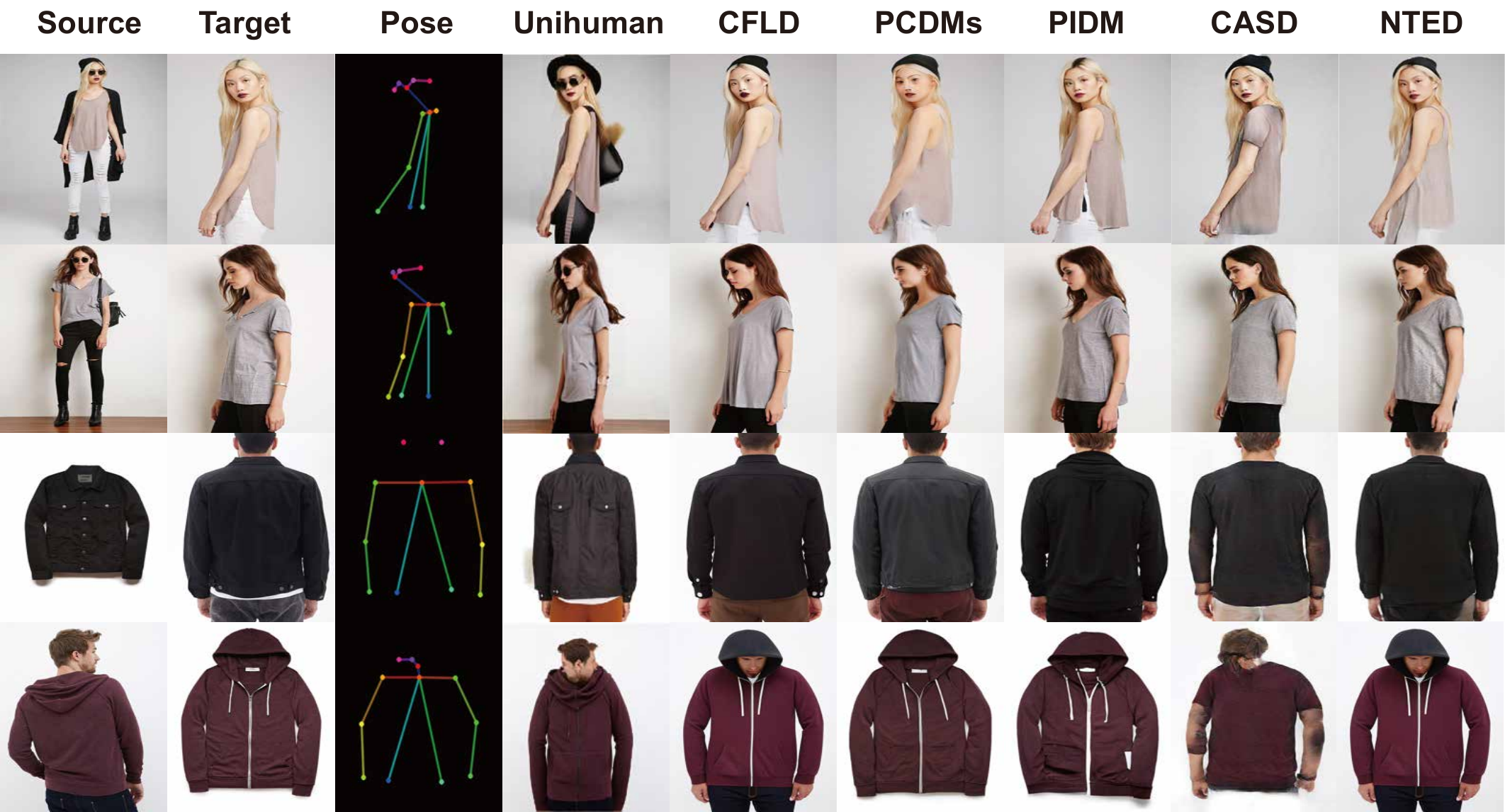}
\caption{Existing methods that are trained on the DeepFashion dataset often exhibit varying degrees of overfitting, despite Unihuman utilizing datasets $\sim$10 times larger than DeepFashion as an additional training dataset. To align with the target image, (row 1) the generated image selectively omits the coat, sunglasses, and other items, (row 2) the background of the generated image alters the background of the source image, (row 3) given only a set of clothing, the model can accurately infer the appearance of a person with their back turned, and (row 4) the model can accurately deduce the clothing.}
\label{fig:overfit}
\end{figure*}

\begin{abstract}
Current Pose-Guided Person Image Synthesis (PGPIS) methods depend heavily on large amounts of labeled triplet data to train the generator in a supervised manner. However, they often falter when applied to in-the-wild samples, primarily due to the distribution gap between the training datasets and real-world test samples. While some researchers aim to enhance model generalizability through sophisticated training procedures, advanced architectures, or by creating more diverse datasets, we adopt the test-time fine-tuning paradigm to customize a pre-trained Text2Image (T2I) model. However, naively applying test-time tuning results in inconsistencies in facial identities and appearance attributes. To address this, we introduce a Visual Consistency Module (VCM), which enhances appearance consistency by combining the face, text, and image embedding. Our approach, named OnePoseTrans, requires only a single source image to generate high-quality pose transfer results, offering greater stability than state-of-the-art data-driven methods. For each test case, OnePoseTrans customizes a model in around 48 seconds with an NVIDIA V100 GPU. Code is available at: \href{https://github.com/Dongqi-Fan/OnePoseTrans}{https://github.com/Dongqi-Fan/OnePoseTrans}
\end{abstract}

\begin{figure*}[htbp]
\centering
\includegraphics[width=\textwidth]{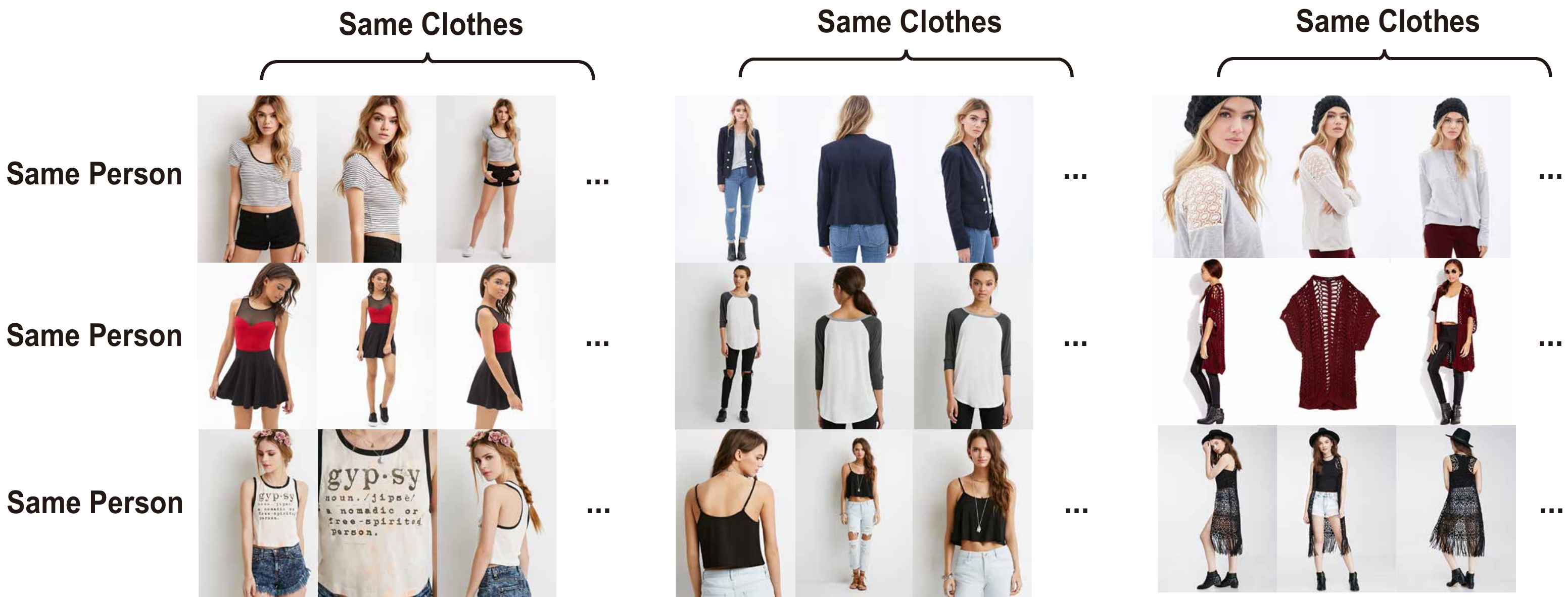}
\caption{The presence of duplicate persons and clothing items and the limited diversity of the studio background within DeepFashion pose a significant challenge for data-driven supervised training.}
\label{fig:duplicate}
\end{figure*}

\section{Introduction}
\label{intro}
Pose-Guided Person Image Synthesis (PGPIS) aims to modify a person's pose in a source image while maintaining the details of both the individual and the background. This task poses a formidable challenge: (1) It inherently involves inpainting, as the model must realistically synthesize portions of the background that might be occluded in the source image; (2) Accurately synthesizing the intricate details and diverse materials of clothing, pants, jewelry, and other accessories, often small and complex in shape, adds significant complexity to the synthesis process; (3) Humans are acutely sensitive to the nuances of the synthesized real-world human image (as opposed to styles such as anime or oil painting). Even subtle imperfections in hairstyle or face shape can lead to a jarring and negative user experience.

Most previous approaches \cite{1unihuman, 2cfld, 3pcdm, 4pidm, 5pocold, 6casd, 7nted, 8dptn} rely on a data-driven supervised learning paradigm using the DeepFashion \cite{11deepfashion} dataset. However, when the test images deviate from the distribution of the training data, artifacts start to appear (See results of WPose and Wild domains in \cref{fig:fig1}). Even if the test images align with the distribution of the training dataset, some unreasonable phenomena still arise (\cref{fig:overfit}). Despite variations in model architecture, the generalizability remains limited due to the lack of diversity in both person and pose in the DeepFashion dataset, which is captured under the controlled studio environments (Examples are shown in \cref{fig:duplicate}).

To address the overfitting problem, Unihuman \cite{1unihuman} introduces the LH-400K dataset, comprising high-quality single-human images selected from LAION-400M \cite{13laion}, with a dataset size 10 times larger than previous collections. Despite these enhancements in diversity and scale, Unihuman's generalizability remains limited when applied to in-the-wild test cases. Another approach \cite{2cfld} seeks to reduce overfitting by introducing a novel Coarse-to-Fine Latent Diffusion method for PGPIS, which employs a progressive training paradigm to decouple the pose and appearance information controls in different stages. However, it does not yield improvements in generalizability, largely due to severe dataset biases.

Unlike previous studies, we adopt a test-time tuning paradigm. Instead of training a model with massive datasets and applying it universally to all test cases, we propose tuning a model individually for each person image. To achieve this, we first use a pre-trained pose-to-image ControlNet \cite{30zhang2023adding} as the control model, which allows the T2I model to be conditioned on a pose image. Then we introduce a novel Visual Consistency Module (VCM) to guarantee the visual attributes consistency, including facial identities, clothing, hairstyle, and accessories. Specifically, we employ a Multimodal Large Language Model (MLLM) to generate textual descriptions of the source person. Next, we incorporate facial style embedding to be integrated into the textual tokens via token replacement. In addition, we segment the source person, encode it into embeddings, and inject these embeddings into the two cross-attention blocks of the UNet.

As the result, OnePoseTrans can successfully customize the base T2I model using only a single source image. Once customized, our model effectively transfers novel poses to the person of interest shown in the source image. The one-shot fine-tuning takes approximately 48 seconds using a single V100 GPU. In summary, our contributions include the following: First, we successfully apply the test-time tuning paradigm to the task of PGPIS; Second, we introduce a novel Visual Consistency Module (VCM) within the model, enhancing the visual consistency of source images. While our one-shot tuning method requires additional fine-tuning time for each person image, it significantly improves generalizability across multiple image domains, particularly for in-the-wild images, as demonstrated by both qualitative and quantitative evaluations.

\begin{figure*}[htbp]
\centering
\includegraphics[width=1\textwidth]{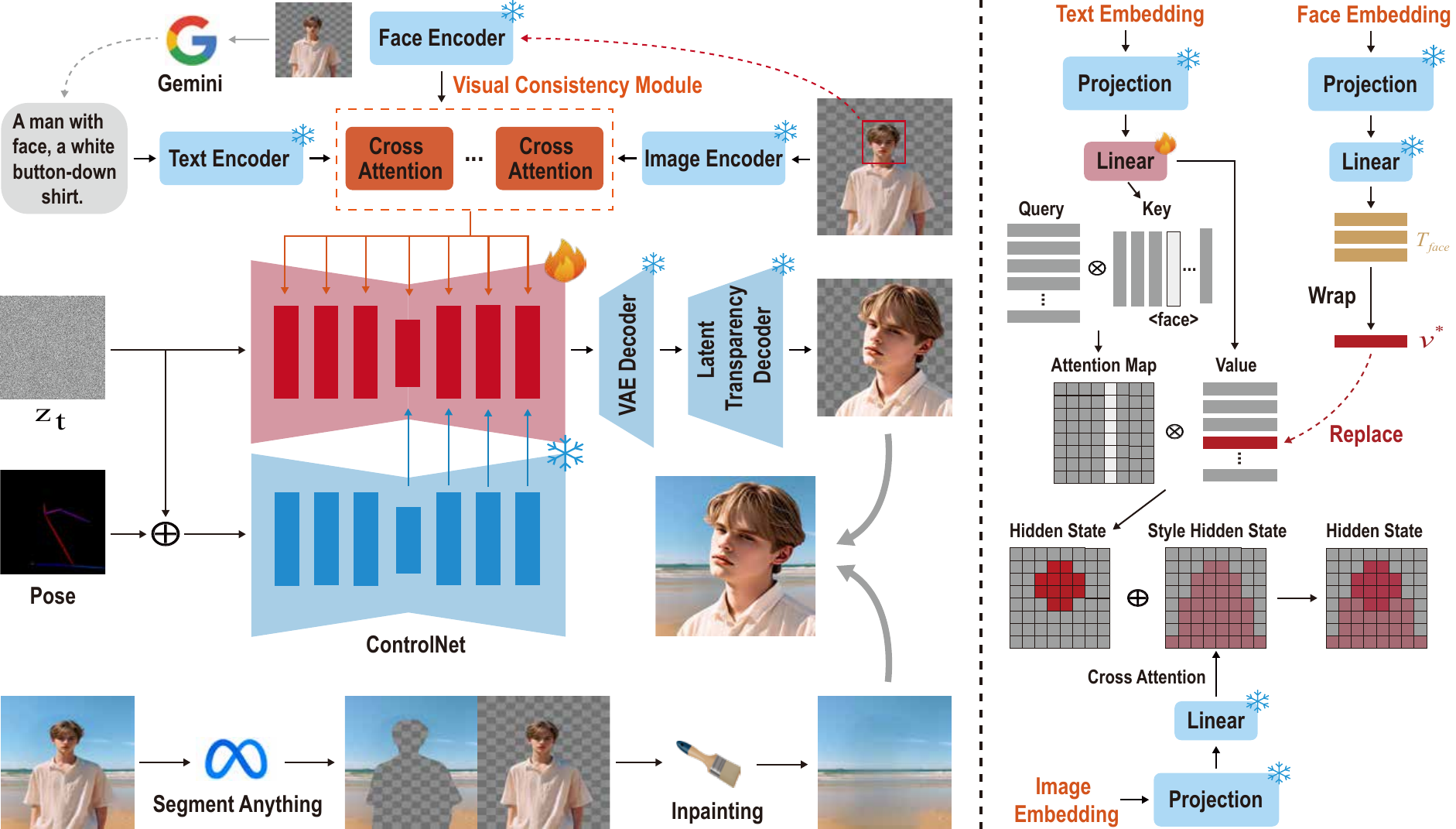}
\caption{(Left) The OnePoseTrans pipeline. During the one-shot tuning stage, only the T2I model, and a portion of the Visual Consistency Module (VCM) are trainable. (Right) The details of the VCM. Note that the $<$face$>$ token replacement is applied exclusively to the value token, without modifying the key and query tokens.}
\label{fig:model}
\end{figure*}

\section{Related Work}
\subsection{Pose-Guided Person Image Synthesis}
To the best of our knowledge, existing diffusion-based approaches for PGPIS can be broadly categorized as supervised learning methods. Because they rely on providing the target image as an additional condition to the U-Net, alongside pose information, to predict the noise for image generation. While these methods differ in their utilization of pose information (e.g., Unihuman \cite{1unihuman} and PoCoLD \cite{5pocold} employ dense pose, while CFLD \cite{2cfld} and PIDM \cite{4pidm} use both open pose images and key points), they share this fundamental reliance on target images during training. This paradigm, however, can lead to overfitting, especially when training data is small and lacks diversity. Consequently, an overfitted model trained on such a dataset might simply “predict” the target image instead of performing image synthesis based on the provided pose. To overcome this limitation, we propose a novel one-shot learning paradigm.

\subsection{Personalizing Generation}
Personalized generation \cite{15gal2022image, 17wei2023elite}, also known as subject-driven generation \cite{16ruiz2023dreambooth, 18chen2023disenbooth, 19chen2024subject}, empowers users to customize image synthesis by providing reference images that embody a desired "concept." This concept, which could be an object, animal, style, human face, or any other visual theme, is learned by the model and used to generate new images that adhere to the concept while involving variations guided by textual prompts. However, accurately capturing and synthesizing a concept from limited reference images, particularly in few-shot, one-shot, or even zero-shot learning scenarios, poses a significant challenge. This difficulty is further amplified when dealing with concepts characterized by complex textures, intricate patterns, or abstract styles. PGPIS can be viewed as a type of personalized generation. However, as discussed in the \cref{intro}, it presents more challenges.

\section{Method}

\subsection{One-Shot Learning}
We leverage the Stable Diffusion XL (SDXL) \cite{23podell2023sdxl} model as our base T2I model, capitalizing on its extensive training on large-scale datasets and its wealth of knowledge. Gemini \cite{24team2023gemini} is utilized to obtain textual descriptions of the source person's appearance. This description encompasses key features such as clothing and accessories, for instance: "A woman with face, a white t-shirt, blue jeans, and a delicate necklace." To eliminate background influence, we employ the Segment Anything model \cite{25kirillov2023segment} to segment out the person as foreground. We then use this segmented foreground, along with the text embedding encoded from the description to efficiently fine-tune SDXL through the LoRA \cite{32hu2021lora} technique:
\begin{equation}
\mathcal{L}=\mathbb{E}_{z_{t}, t, I_{source}, P_{text}, \epsilon \sim \mathcal{N}(0,1)}{\| \epsilon-\epsilon_{\theta}(z_{t}, t, I_{source}, P_{text}) \|}_{2}^{2}
\end{equation}

Where t is the timestep, $z_{t}$ is the random Gaussian noise, $I_{source}$ is the foreground person, $P_{text}$ is the given text and $\epsilon$ represent latent noise at timestep t. $\epsilon_{\theta}$ represents U-Net in SDXL, which receive $z_{t}$, t, $I_{source}$, $P_{text}$ and output predicted noise $\hat{\epsilon}$.

\begin{figure*}[htbp]
\centering
\includegraphics[width=0.95\textwidth]{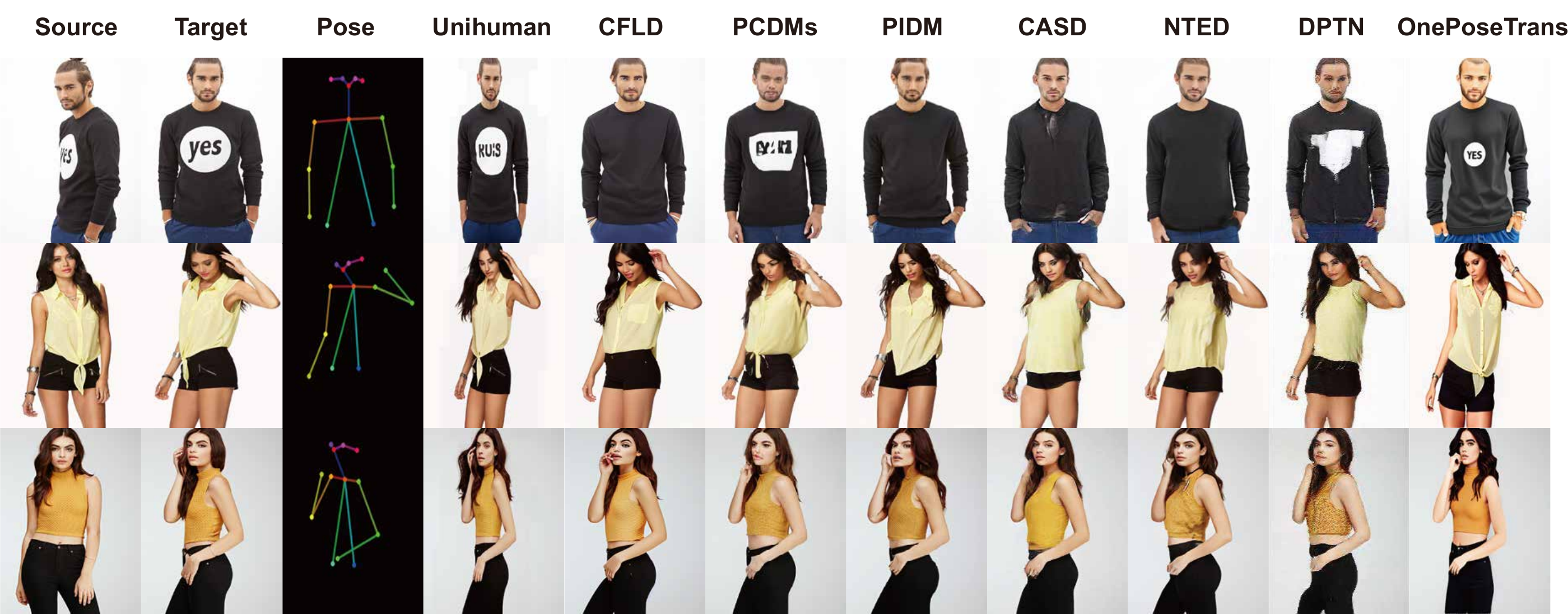}
\caption{Visual comparisons with state-of-the-art models on the DeepFashion dataset. Our OnePoseTrans achieves comparable visual results to these models.}
\label{visual_fashion}
\end{figure*}

\begin{figure*}[htbp]
\centering
\includegraphics[width=0.75\textwidth]{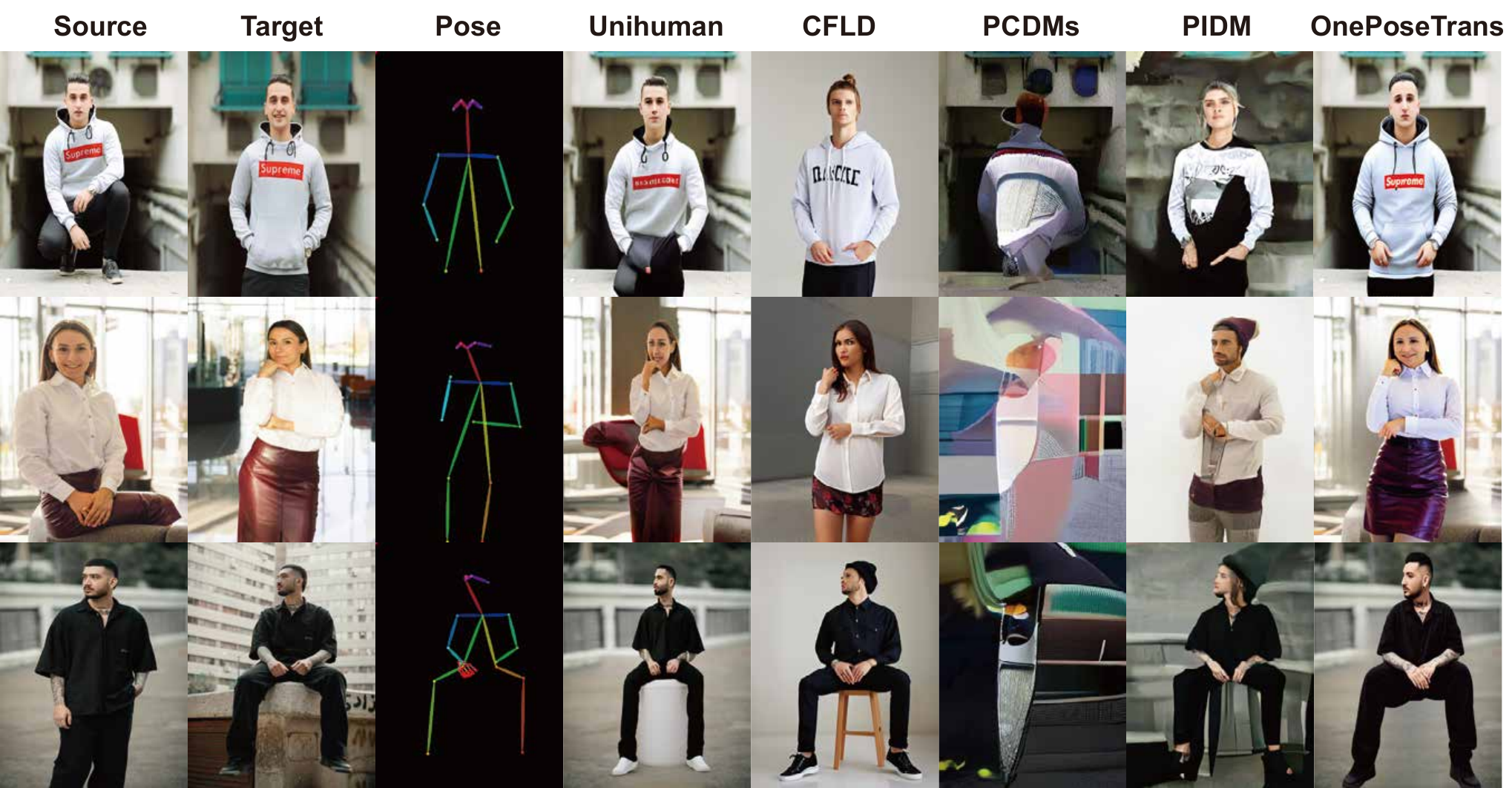}
\caption{Visual comparisons with state-of-the-art models on the WPose dataset. Our OnePoseTrans achieves best visual quality.}
\label{visual_wpose}
\end{figure*}

\subsection{Pose-Guided Person Image Synthesis}
After one-shot training ends, the whole VCM, ControlNet \cite{30zhang2023adding}, VAE Decoder, and Latent Transparency Decoder \cite{26zhang2024transparent} are incorporated to perform PGPIS. The Latent Transparency Decoder guarantees that the output is an RGBA image. A single step in the diffusion process can be represented as:
\begin{equation}
\epsilon_{\theta}(z_{t}, t, I_{pose}, I_{source}, P_{text}, P_{face}) \rightarrow \hat{\epsilon}
\end{equation}

Where $I_{pose}$ and $P_{face}$ represent the pose image and face embedding encoded from the source image. The tuned T2I model processes  $z_{t}$, t, $P_{text}$, $I_{source}$, $I_{pose}$ and $P_{face}$ as inputs, producing the new foreground after t timesteps. Specifically, the diffusion process in OnePoseTrans involves style injection, LoRA weight injection, and layer-diffuse weight injection, which are elaborated as below.

\begin{figure*}[htbp]
\centering
\includegraphics[width=\textwidth]{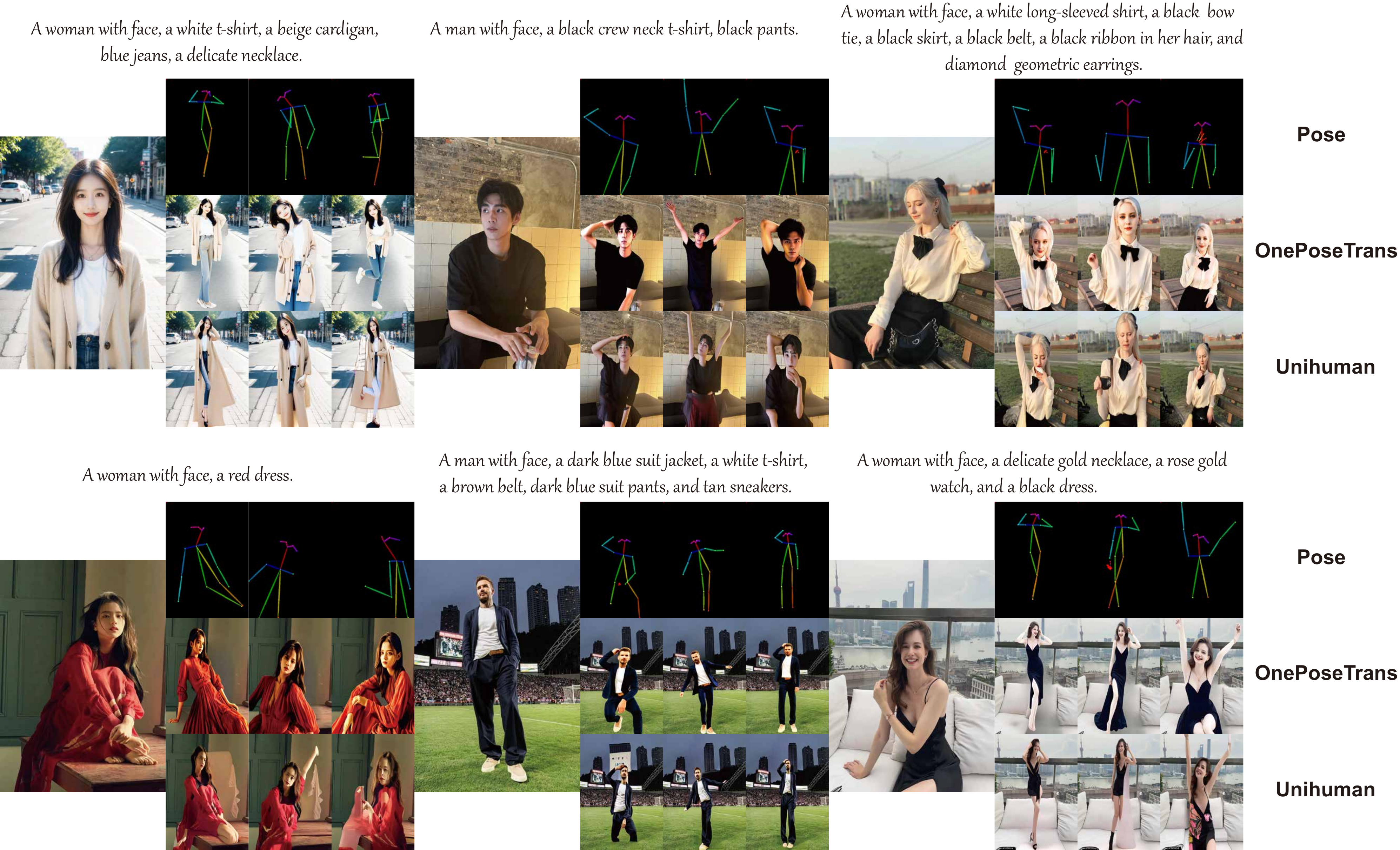}
\caption{Visual comparisons between OnePose and Unihuman \cite{1unihuman} in the wild. }
\label{visual_wild}
\end{figure*}

\subsubsection{Style Injection}
While the IP-Adapter \cite{14ipadapter} introduces additional cross-attention blocks within the U-Net to achieve style injection by adding computed style hidden states, InstantStyle \cite{27wang2024instantstyle} observes that different attention layers capture style information differently. Specifically, they find up.blocks.0.attentions.1 capture style and down.blocks.2.attentions.1 capture spatial layout. However, our analysis arrives at two different insights: (1) up.blocks.0.attentions.0 also contributes to style capture; (2) the presence of other attention layers will harm the PGPIS. Therefore, in our VCM, we exclusively use image embedding and style hidden states in up.blocks.0.attentions.0 and up.blocks.0.attentions.1.

\subsubsection{Weight Injection}
\textit{LoRA Weight Injection:} We further explore the influence of LoRA's weight injection on the various attention blocks. Our experiment reveals that only up.blocks.0.attentions.1 significantly contributes to PGPIS, while the remaining attention blocks are beneficial only when the injection intensity is low. Therefore, we set the scale of up.blocks.0.attentions.1 to 1 and the scale of the other attention blocks to 0.2. \textit{LayerDiffuse Weight Injection}: LayerDiffuse \cite{26zhang2024transparent} is a method that enables large-scale pre-trained latent diffusion models to generate transparent images. LayerDiffuse has a trained plug-and-play weight offset and a Latent Transparency Decoder. Assuming the original weight of the base diffusion model is $w$ and the weight offset $w'$, the new weight of the model can be expressed as $w = w + w'$.

\subsubsection{Fine-Grained Identity-Preserving}
Face identity-preserving generation aims to ensure the generated face accurately reflects the identity of a face provided in user-specified images. Recently, there have been many excellent works in this area. For instance, FasteComposer \cite{28xiao2023fastcomposer} and PhotoMaker \cite{21li2024photomaker} wrap the ID embeddings of reference images into special words within the text prompt, such as 'woman' or 'man'. On the other hand, InstantID \cite{20wang2024instantid}, similar to IP-Adapter-Face \cite{14ipadapter}, integrates the hidden states of facial styles with the hidden states of cross-attention to achieve personalized generation.  However, existing methods often struggle with preserving face consistency in PGPIS. This issue arises because facial features, acting as a stylistic element, can clash with other concurrent styles.

To overcome this challenge, we propose a novel fine-grained identity-preserving method that focuses on editing the facial region while minimizing interference with other stylistic elements. Our approach begins with extracting a source face embedding using a face detection model\footnote{https://github.com/deepinsight/insightface}. As illustrated in Fig.\ref{fig:model} right, this embedding is then processed using a pre-trained face Projection and Linear module borrowed from the InstantID \cite{20wang2024instantid}. This process yields face tokens $T_{face}$ which are then wrapped into a single token $v^{*}$ and used to replace the corresponding $<$face$>$ token in the value:

\begin{equation}
T_{face}=(T_{1} \dots T_{N})=Linear(Projection(P_{face}))
\end{equation}

\begin{equation}
v^{*}=s\cdot\sum_{i=1}^{N}softmax({T_{face}})^{(i)}T_{face}^{(i)}
\end{equation}

Where s is the fixed control constant, the higher s is, the greater the editing intensity. In this paper, we fixed s to 4.

\begin{table*}[htbp]
\caption{Quantitative comparison of OnePoseTrans with other state-of-the-art models on the DeepFashion (top) and WPose (bottom) datasets. There's a significant performance gap on DeepFashion, which can be largely attributed to severe overfitting issues in some existing data-driven methods.}
\centering
\scalebox{0.9}{
\renewcommand{\arraystretch}{1.1}
\begin{tabular}{cccccc}
\toprule[1.5pt]
Dataset                      & Method   & PSNR $\uparrow$  & SSIM $\uparrow$  & LPIPS $\downarrow$ & DINO $\uparrow$  \\ \midrule[0.9pt]
\multirow{9}{*}{DeepFashion \cite{11deepfashion}} & Unihuman \cite{1unihuman} & 13.31 & 0.5865 & 0.3365 & 0.9434 \\
                             & CFLD \cite{2cfld}     & 17.65 & \textbf{0.7104} & 0.2040 & 0.9731 \\
                             & PCDMs \cite{3pcdm}    & \textbf{17.81} & 0.7022 & \textbf{0.1982} & 0.9720 \\
                             & PIDM \cite{4pidm}     & 17.15 & 0.6904 & 0.2048 & \textbf{0.9754} \\
                             & CASD \cite{6casd}     & 17.36 & 0.7018 & 0.2222 & 0.9598 \\
                             & NTED \cite{7nted}     & 17.33 & 0.6933 & 0.2207 & 0.9585 \\
                             & DPTN \cite{8dptn}     & 17.36 & 0.6837 & 0.2089 & 0.9490 \\
                             & \textbf{OnePose}  & 13.20 & 0.5998 & 0.3119 & 0.9421 \\
                             & \textbf{OnePose} $\dag$ & 13.57 & 0.6053 & 0.3066 & 0.9476 \\
                             \midrule[0.9pt]
\multirow{6}{*}{WPose \cite{1unihuman}}       & Unihuman \cite{1unihuman} & \textbf{10.67} & 0.2038 & 0.5511 & 0.8324 \\
                             & CFLD \cite{2cfld}     & 9.67  & 0.2760 & 0.6170 & 0.7627 \\
                             & PCDMs \cite{3pcdm}    & 9.68  & 0.1886 & 0.6626 & 0.6389 \\
                             & PIDMc\cite{4pidm}     & 9.82  & 0.2212 & 0.6050 & 0.7168 \\
                             & \textbf{OnePose}   & 10.27 & 0.2170 & 0.5527 & 0.8404 \\
                             & \textbf{OnePose} $\dag$ & 10.46 & \textbf{0.2212} & \textbf{0.5510} & \textbf{0.8401} \\ \bottomrule[1.5pt]
\end{tabular}}
\label{table1}
\end{table*}

\begin{table}[htbp]
\caption{Ablation results with other state-of-the-art face identity-preserving techniques on the DeepFashion dataset.}
\vspace{3mm}
\centering
\scalebox{0.78}{
\renewcommand{\arraystretch}{1.15}
\begin{tabular}{ccccc}
\toprule[1.5pt]
Method   & PSNR $\uparrow$  & SSIM $\uparrow$  & LPIPS $\downarrow$ & DINO $\uparrow$ \\ \midrule[0.9pt] 
\begin{tabular}[c]{@{}c@{}}OnePoseTrans +\\[-0.3ex] VCM (Ours)\end{tabular}   & 13.20 & 0.5998 & 0.3119 & 0.9421 \\
\begin{tabular}[c]{@{}c@{}}OnePoseTrans $\dag$ +\\[-0.3ex] VCM (Ours)\end{tabular}   & \textbf{13.57} & \textbf{0.6053} & \textbf{0.3066} & \textbf{0.9476} \\
\begin{tabular}[c]{@{}c@{}}OnePoseTrans +\\[-0.3ex] Full Token Replacement \cite{21li2024photomaker}\end{tabular}  & 12.65 & 0.5992 & 0.3431 & 0.9296 \\
\begin{tabular}[c]{@{}c@{}}OnePoseTrans +\\[-0.3ex] Cross-Attention \cite{20wang2024instantid}\end{tabular}   & 13.00 & 0.5980 & 0.3235 & 0.9388 \\ 
                             \bottomrule[1.5pt]
\end{tabular}}
\label{table2}
\end{table}

\subsection{Consistency in Foreground and Background}
The synthesized foreground images generated by OnePoseTrans occasionally exhibit inconsistencies in color and brightness when compared to the background of the source images. To mitigate this issue, we introduce a refining model subsequent to OnePoseTrans to enhance foreground-background consistency. This refining model is derived from the original OnePoseTrans paradigm by removing the ControlNet module and incorporating a Latent Transparency Encoder. We denote this enhanced OnePoseTrans with the refining model as \textbf{OnePoseTrans $\dag$}. Note that OnePoseTrans alone is generally capable of effectively performing PGPIS, and the use of \textbf{OnePoseTrans $\dag$} is not universally required.

\section{Experiment}

\subsection{Implementation details}
During the one-shot learning stage, we employ LoRA technology to fine-tune the attention layers within the SDXL U-Net. The LoRA rank is set to 32, with 60 training iterations, a batch size of 2, and a learning rate of 0.001. All experiments present in this paper are conducted on an NVIDIA V100 GPU, with a fine-tuning time of approximately 48 seconds. For evaluation, we conduct quantitative experiments on the complete WPose dataset and a subset of 1,763 sample pairs from the DeepFashion test dataset. This subset is carefully selected due to the presence of numerous unreasonable sample pairs within the original DeepFashion test set (see \cref{fig:overfit} for examples), exhibiting inconsistencies in clothing and background. As our OnePoseTrans utilizes a single source image, generating images consistent with such unreasonable target images is not feasible. To ensure a fair comparison, our evaluations of DeepFashion against other methods are restricted to this curated subset of 1,763 sample pairs.

\subsection{Compare with Other Works}
\cref{table1} shows the evaluation results on the DeepFashion and WPose datasets. On the DeepFashion dataset, OnePoseTrans and Unihuman show lower performance compared to other models, as the latter were trained on the DeepFashion training split, which follows a similar distribution to the test data. On the more challenging WPose dataset, OnePoseTrans and Unihuman outperform other models by a significant margin. OnePoseTrans achieves PSNR scores comparable to Unihuman while surpassing it in SSIM, LPIPS, and DINO. Visual comparisons on the DeepFashion domain, WPose domain, and Wild domain are provided in \cref{visual_fashion}, \cref{visual_wpose}, and \cref{visual_wild}, respectively.

\subsection{Ablation Studies}
\begin{figure}[t]
\centering
\includegraphics[width=0.48\textwidth]{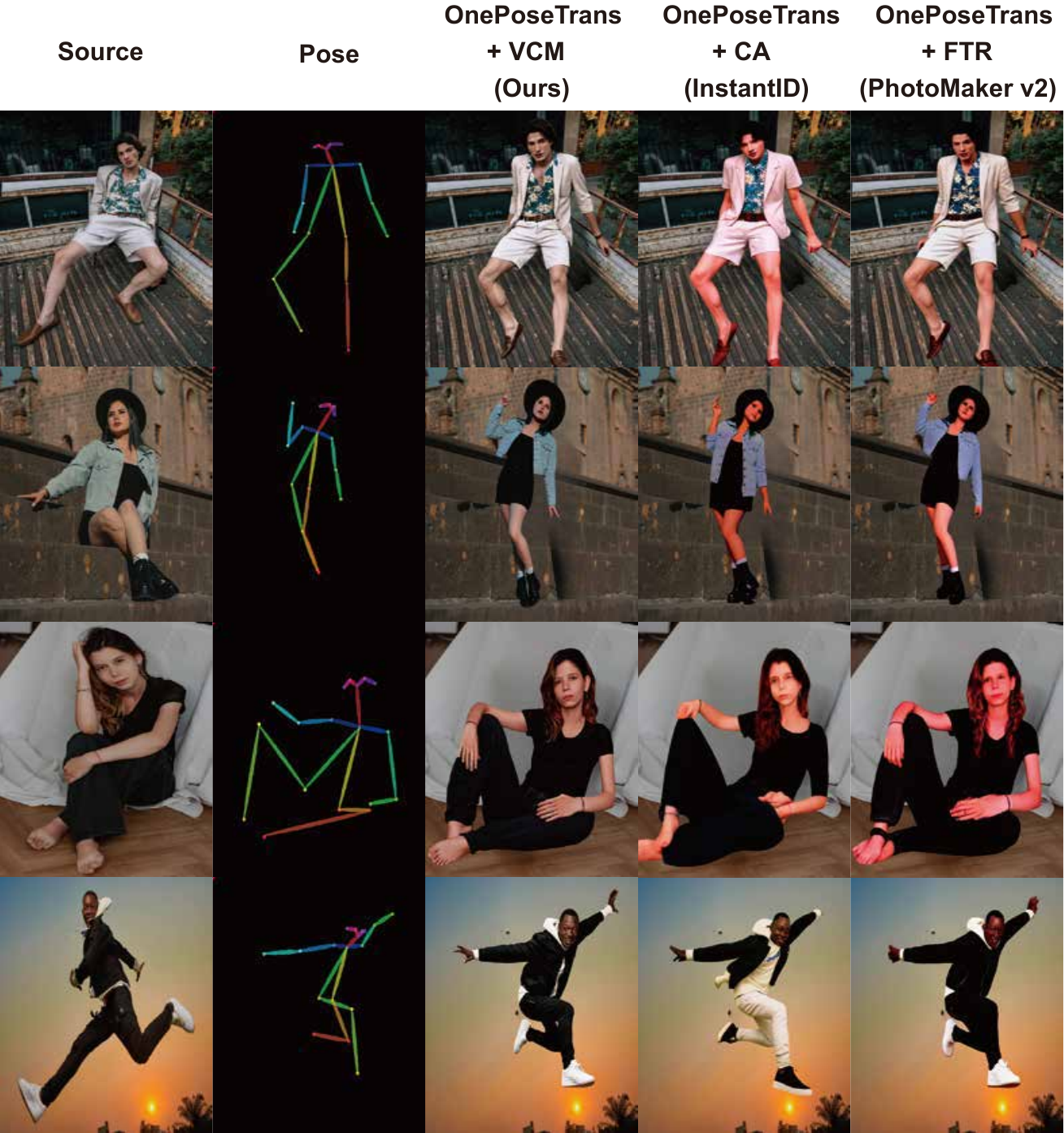}
\caption{Visual comparisons with other state-of-the-art face identity-preserving techniques on the WPose dataset.}
\label{visual_ablation}
\end{figure}

To assess the effectiveness of the Visual Consistency Module (VCM) in preserving facial identities, we conduct a comparative experiment against two state-of-the-art face identity-preserving techniques: the traditional full token replacement technique from Photomaker v2 \cite{21li2024photomaker} and the cross-attention technique from InstantID \cite{20wang2024instantid}. We replace our face identity-preserving module with each of these techniques, keeping all other components of the pipeline unchanged. As shown in \cref{table2}, our method significantly outperforms both Photomaker v2 and InstantID, underscoring the superiority of our Visual Consistency Module in preserving facial identities. Visual comparasion are provided in \cref{visual_ablation}.

\section{Conclusion}
For the first time, we introduce the test-time tuning paradigm to the pose transfer task. To tackle visual inconsistencies, we propose a novel Visual Consistency Module that enhances the consistency of facial identities and appearance attributes in source images. Our OnePoseTrans method requires only a single image to fine-tune the T2I model, enabling effective transfer of novel poses to the source person. OnePoseTrans demonstrates superior generalization across multiple domains compared to existing data-driven methods.

\section{Limitation and Future Work}
Despite the effectiveness of OnePoseTrans in mitigating overfitting through its novel paradigm and its demonstrated robust generalization capabilities, certain limitations warrant further investigation: (1) Inpainting Model: Our experiments with various state-of-the-art inpainting models, both diffusion-based and non-diffusion-based, reveal limitations. Diffusion-based models often introduced extraneous objects or persons, while non-diffusion-based models frequently yielded subpar results. (2) Description Generation: The process of crafting accurate and consistent descriptions requires meticulous attention. Automated batch generation using Gemini for the WPose and DeepFashion datasets results in instances of inaccuracies, errors, and deviations from the defined template. These low-quality descriptions negatively impact PGPIS performance, and manual inspection and correction of each description is deemed impractical. (3) Foreground-Background Consistency: While the additional refining model helps alleviate inconsistencies between the synthesized foreground and background, subtle discrepancies, such as variations in brightness levels, may still arise. Future research will prioritize addressing these limitations. This includes exploring alternative inpainting strategies, refining the description generation process, and further enhancing foreground-background consistency. Ultimately, our goal is to improve OnePoseTrans's overall performance, streamline the pipeline for increased automation, and reduce inference time.

{
    \small
    \bibliographystyle{ieeenat_fullname}
    \bibliography{main}

\begin{thebibliography}{26}
\providecommand{\natexlab}[1]{#1}
\providecommand{\url}[1]{\texttt{#1}}
\expandafter\ifx\csname urlstyle\endcsname\relax
  \providecommand{\doi}[1]{doi: #1}\else
  \providecommand{\doi}{doi: \begingroup \urlstyle{rm}\Url}\fi

\bibitem[Bhunia et~al.(2023)Bhunia, Khan, Cholakkal, Anwer, Laaksonen, Shah, and Khan]{4pidm}
Ankan~Kumar Bhunia, Salman Khan, Hisham Cholakkal, Rao~Muhammad Anwer, Jorma Laaksonen, Mubarak Shah, and Fahad~Shahbaz Khan.
\newblock Person image synthesis via denoising diffusion model.
\newblock In \emph{Proceedings of the IEEE/CVF Conference on Computer Vision and Pattern Recognition}, pages 5968--5976, 2023.

\bibitem[Chen et~al.(2023)Chen, Zhang, Wu, Wang, Duan, Zhou, and Zhu]{18chen2023disenbooth}
Hong Chen, Yipeng Zhang, Simin Wu, Xin Wang, Xuguang Duan, Yuwei Zhou, and Wenwu Zhu.
\newblock Disenbooth: Identity-preserving disentangled tuning for subject-driven text-to-image generation.
\newblock \emph{arXiv preprint arXiv:2305.03374}, 2023.

\bibitem[Chen et~al.(2024)Chen, Hu, Li, Ruiz, Jia, Chang, and Cohen]{19chen2024subject}
Wenhu Chen, Hexiang Hu, Yandong Li, Nataniel Ruiz, Xuhui Jia, Ming-Wei Chang, and William~W Cohen.
\newblock Subject-driven text-to-image generation via apprenticeship learning.
\newblock \emph{Advances in Neural Information Processing Systems}, 36, 2024.

\bibitem[Gal et~al.(2022)Gal, Alaluf, Atzmon, Patashnik, Bermano, Chechik, and Cohen-Or]{15gal2022image}
Rinon Gal, Yuval Alaluf, Yuval Atzmon, Or Patashnik, Amit~H Bermano, Gal Chechik, and Daniel Cohen-Or.
\newblock An image is worth one word: Personalizing text-to-image generation using textual inversion.
\newblock \emph{arXiv preprint arXiv:2208.01618}, 2022.

\bibitem[Han et~al.(2023)Han, Zhu, Deng, Song, and Xiang]{5pocold}
Xiao Han, Xiatian Zhu, Jiankang Deng, Yi-Zhe Song, and Tao Xiang.
\newblock Controllable person image synthesis with pose-constrained latent diffusion.
\newblock In \emph{Proceedings of the IEEE/CVF International Conference on Computer Vision}, pages 22768--22777, 2023.

\bibitem[Hu et~al.(2021)Hu, Shen, Wallis, Allen-Zhu, Li, Wang, Wang, and Chen]{32hu2021lora}
Edward~J Hu, Yelong Shen, Phillip Wallis, Zeyuan Allen-Zhu, Yuanzhi Li, Shean Wang, Lu Wang, and Weizhu Chen.
\newblock Lora: Low-rank adaptation of large language models.
\newblock \emph{arXiv preprint arXiv:2106.09685}, 2021.

\bibitem[Kirillov et~al.(2023)Kirillov, Mintun, Ravi, Mao, Rolland, Gustafson, Xiao, Whitehead, Berg, Lo, et~al.]{25kirillov2023segment}
Alexander Kirillov, Eric Mintun, Nikhila Ravi, Hanzi Mao, Chloe Rolland, Laura Gustafson, Tete Xiao, Spencer Whitehead, Alexander~C Berg, Wan-Yen Lo, et~al.
\newblock Segment anything.
\newblock In \emph{Proceedings of the IEEE/CVF International Conference on Computer Vision}, pages 4015--4026, 2023.

\bibitem[Li et~al.(2024{\natexlab{a}})Li, Liu, Singh, Wang, Zhang, Plummer, and Lin]{1unihuman}
Nannan Li, Qing Liu, Krishna~Kumar Singh, Yilin Wang, Jianming Zhang, Bryan~A Plummer, and Zhe Lin.
\newblock Unihuman: A unified model for editing human images in the wild.
\newblock In \emph{Proceedings of the IEEE/CVF Conference on Computer Vision and Pattern Recognition}, pages 2039--2048, 2024{\natexlab{a}}.

\bibitem[Li et~al.(2024{\natexlab{b}})Li, Cao, Wang, Qi, Cheng, and Shan]{21li2024photomaker}
Zhen Li, Mingdeng Cao, Xintao Wang, Zhongang Qi, Ming-Ming Cheng, and Ying Shan.
\newblock Photomaker: Customizing realistic human photos via stacked id embedding.
\newblock In \emph{Proceedings of the IEEE/CVF Conference on Computer Vision and Pattern Recognition}, pages 8640--8650, 2024{\natexlab{b}}.

\bibitem[Liu et~al.(2016)Liu, Luo, Qiu, Wang, and Tang]{11deepfashion}
Ziwei Liu, Ping Luo, Shi Qiu, Xiaogang Wang, and Xiaoou Tang.
\newblock Deepfashion: Powering robust clothes recognition and retrieval with rich annotations.
\newblock In \emph{Proceedings of the IEEE conference on computer vision and pattern recognition}, pages 1096--1104, 2016.

\bibitem[Lu et~al.(2024)Lu, Zhang, Ma, Xie, and Lai]{2cfld}
Yanzuo Lu, Manlin Zhang, Andy~J Ma, Xiaohua Xie, and Jianhuang Lai.
\newblock Coarse-to-fine latent diffusion for pose-guided person image synthesis.
\newblock In \emph{Proceedings of the IEEE/CVF Conference on Computer Vision and Pattern Recognition}, pages 6420--6429, 2024.

\bibitem[Podell et~al.(2023)Podell, English, Lacey, Blattmann, Dockhorn, M{\"u}ller, Penna, and Rombach]{23podell2023sdxl}
Dustin Podell, Zion English, Kyle Lacey, Andreas Blattmann, Tim Dockhorn, Jonas M{\"u}ller, Joe Penna, and Robin Rombach.
\newblock Sdxl: Improving latent diffusion models for high-resolution image synthesis.
\newblock \emph{arXiv preprint arXiv:2307.01952}, 2023.

\bibitem[Ren et~al.(2022)Ren, Fan, Li, Liu, and Li]{7nted}
Yurui Ren, Xiaoqing Fan, Ge Li, Shan Liu, and Thomas~H Li.
\newblock Neural texture extraction and distribution for controllable person image synthesis.
\newblock In \emph{Proceedings of the IEEE/CVF conference on computer vision and pattern recognition}, pages 13535--13544, 2022.

\bibitem[Ruiz et~al.(2023)Ruiz, Li, Jampani, Pritch, Rubinstein, and Aberman]{16ruiz2023dreambooth}
Nataniel Ruiz, Yuanzhen Li, Varun Jampani, Yael Pritch, Michael Rubinstein, and Kfir Aberman.
\newblock Dreambooth: Fine tuning text-to-image diffusion models for subject-driven generation.
\newblock In \emph{Proceedings of the IEEE/CVF conference on computer vision and pattern recognition}, pages 22500--22510, 2023.

\bibitem[Schuhmann et~al.(2021)Schuhmann, Vencu, Beaumont, Kaczmarczyk, Mullis, Katta, Coombes, Jitsev, and Komatsuzaki]{13laion}
Christoph Schuhmann, Richard Vencu, Romain Beaumont, Robert Kaczmarczyk, Clayton Mullis, Aarush Katta, Theo Coombes, Jenia Jitsev, and Aran Komatsuzaki.
\newblock Laion-400m: Open dataset of clip-filtered 400 million image-text pairs.
\newblock \emph{arXiv preprint arXiv:2111.02114}, 2021.

\bibitem[Shen et~al.(2023)Shen, Ye, Zhang, Wang, Han, and Yang]{3pcdm}
Fei Shen, Hu Ye, Jun Zhang, Cong Wang, Xiao Han, and Wei Yang.
\newblock Advancing pose-guided image synthesis with progressive conditional diffusion models.
\newblock \emph{arXiv preprint arXiv:2310.06313}, 2023.

\bibitem[Team et~al.(2023)Team, Anil, Borgeaud, Wu, Alayrac, Yu, Soricut, Schalkwyk, Dai, Hauth, et~al.]{24team2023gemini}
Gemini Team, Rohan Anil, Sebastian Borgeaud, Yonghui Wu, Jean-Baptiste Alayrac, Jiahui Yu, Radu Soricut, Johan Schalkwyk, Andrew~M Dai, Anja Hauth, et~al.
\newblock Gemini: a family of highly capable multimodal models.
\newblock \emph{arXiv preprint arXiv:2312.11805}, 2023.

\bibitem[Wang et~al.(2024{\natexlab{a}})Wang, Wang, Bai, Qin, and Chen]{27wang2024instantstyle}
Haofan Wang, Qixun Wang, Xu Bai, Zekui Qin, and Anthony Chen.
\newblock Instantstyle: Free lunch towards style-preserving in text-to-image generation.
\newblock \emph{arXiv preprint arXiv:2404.02733}, 2024{\natexlab{a}}.

\bibitem[Wang et~al.(2024{\natexlab{b}})Wang, Bai, Wang, Qin, and Chen]{20wang2024instantid}
Qixun Wang, Xu Bai, Haofan Wang, Zekui Qin, and Anthony Chen.
\newblock Instantid: Zero-shot identity-preserving generation in seconds.
\newblock \emph{arXiv preprint arXiv:2401.07519}, 2024{\natexlab{b}}.

\bibitem[Wei et~al.(2023)Wei, Zhang, Ji, Bai, Zhang, and Zuo]{17wei2023elite}
Yuxiang Wei, Yabo Zhang, Zhilong Ji, Jinfeng Bai, Lei Zhang, and Wangmeng Zuo.
\newblock Elite: Encoding visual concepts into textual embeddings for customized text-to-image generation.
\newblock In \emph{Proceedings of the IEEE/CVF International Conference on Computer Vision}, pages 15943--15953, 2023.

\bibitem[Xiao et~al.(2023)Xiao, Yin, Freeman, Durand, and Han]{28xiao2023fastcomposer}
Guangxuan Xiao, Tianwei Yin, William~T Freeman, Fr{\'e}do Durand, and Song Han.
\newblock Fastcomposer: Tuning-free multi-subject image generation with localized attention.
\newblock \emph{arXiv preprint arXiv:2305.10431}, 2023.

\bibitem[Ye et~al.(2023)Ye, Zhang, Liu, Han, and Yang]{14ipadapter}
Hu Ye, Jun Zhang, Sibo Liu, Xiao Han, and Wei Yang.
\newblock Ip-adapter: Text compatible image prompt adapter for text-to-image diffusion models.
\newblock \emph{arXiv preprint arXiv:2308.06721}, 2023.

\bibitem[Zhang and Agrawala(2024)]{26zhang2024transparent}
Lvmin Zhang and Maneesh Agrawala.
\newblock Transparent image layer diffusion using latent transparency.
\newblock \emph{arXiv preprint arXiv:2402.17113}, 2024.

\bibitem[Zhang et~al.(2023)Zhang, Rao, and Agrawala]{30zhang2023adding}
Lvmin Zhang, Anyi Rao, and Maneesh Agrawala.
\newblock Adding conditional control to text-to-image diffusion models.
\newblock In \emph{Proceedings of the IEEE/CVF International Conference on Computer Vision}, pages 3836--3847, 2023.

\bibitem[Zhang et~al.(2022)Zhang, Yang, Lai, and Xie]{8dptn}
Pengze Zhang, Lingxiao Yang, Jian-Huang Lai, and Xiaohua Xie.
\newblock Exploring dual-task correlation for pose guided person image generation.
\newblock In \emph{Proceedings of the IEEE/CVF Conference on Computer Vision and Pattern Recognition}, pages 7713--7722, 2022.

\bibitem[Zhou et~al.(2022)Zhou, Yin, Chen, Sun, Gao, and Li]{6casd}
Xinyue Zhou, Mingyu Yin, Xinyuan Chen, Li Sun, Changxin Gao, and Qingli Li.
\newblock Cross attention based style distribution for controllable person image synthesis.
\newblock In \emph{European Conference on Computer Vision}, pages 161--178. Springer, 2022.

\end{thebibliography}
}


\end{document}